# Exponential Regret Bounds for Gaussian Process Bandits with Deterministic Observations

**Nando de Freitas**  NANDO@CS.UBC.CA
Department of Computer Science, University of British Columbia, Vancouver, BC V6T 1Z4, Canada

**Alex J. Smola**  ALEX@SMOLA.ORG
Yahoo! Research, Santa Clara, CA 95051, USA

**Masrour Zoghi**  MZOGHI@CS.UBC.CA
Department of Computer Science, University of British Columbia, Vancouver, BC V6T 1Z4, Canada

## Abstract

This paper analyzes the problem of Gaussian process (GP) bandits with deterministic observations. The analysis uses a branch and bound algorithm that is related to the UCB algorithm of (Srinivas et al., 2010). For GPs with Gaussian observation noise, with variance strictly greater than zero, (Srinivas et al., 2010) proved that the regret vanishes at the approximate rate of $\mathcal{O}\left(\frac{1}{\sqrt{t}}\right)$, where $t$ is the number of observations. To complement their result, we attack the deterministic case and attain a much faster exponential convergence rate. Under some regularity assumptions, we show that the regret decreases asymptotically according to $\mathcal{O}\left(e^{-\frac{\tau t}{(\ln t)^{d/4}}}\right)$ with high probability. Here, $d$ is the dimension of the search space and $\tau$ is a constant that depends on the behaviour of the objective function near its global maximum.

## 1. Introduction

Let $f : \mathcal{D} \to \mathbb{R}$ be a function on a compact subset $\mathcal{D} \subseteq \mathbb{R}^d$. We would like to address the global optimization problem

$$x_M = \underset{x \in \mathcal{D}}{\operatorname{argmax}} f(x).$$

Let us assume for the sake of simplicity that the objective function $f$ has a unique global maximum (although it may have many local maxima).

The space $\mathcal{D}$ might be the set of free parameters that one could feed into a time-consuming algorithm or the locations where a sensor could be deployed, and the function $f$ might be a measure of the performance of the algorithm (e.g. how long it takes to run). We refer the reader to (Močkus, 1982; Schonlau et al., 1998; Gramacy et al., 2004; Brochu et al., 2007; Lizotte, 2008; Martinez–Cantin et al., 2009; Garnett et al., 2010) for practical examples. In this paper, our assumption is that once the function has been probed at point $x \in \mathcal{D}$, then the value $f(x)$ can be observed with very high precision. This is the case when the deployed sensors are very accurate or if the algorithm is deterministic. An example of this is the configuration of CPLEX parameters in mixed-integer programming (Hutter et al., 2010). More ambitiously, we might be interested in the *simultaneous* automatic configuration of an entire system (algorithms, architectures and hardware) whose performance is deterministic in terms of several free parameters and design choices.

Global optimization is a difficult problem without any assumptions on the objective function $f$. The main complicating factor is the uncertainty over the extent of the variations of $f$, e.g. one could consider the characteristic function, which is equal to 1 at $x_M$ and 0 elsewhere, and none of the methods we mention here can optimize this function without exhaustively searching through every point in $\mathcal{D}$.

The way a large number of global optimization methods address this problem is by imposing some prior assumption on how fast the objective function $f$ can vary. The most explicit manifestation of this remedy is the imposition of a Lipschitz assumption on $f$, which requires the change in the value of $f(x)$, as the point $x$ moves around, to be smaller than a constant multiple of the distance traveled by $x$ (Hansen et al., 1992). As pointed out in (Bubeck et al., 2011, Figure 3), it is only important to have this kind of tight control over the function near its optimum: elsewhere in the space, we can have what they have dubbed a "weak Lipschitz" condition.

One way to relax these hard Lipschitz constraints is by putting a Gaussian Process (GP) prior on the function. Instead of restricting the function from oscillating too fast, a GP prior requires those fast oscillations to have low probability, cf. (Ghosal & Roy, 2006, Theorem 5).

The main point of these bounds (be they hard or soft) is to assist with the *exploration-exploitation trade-off* that global optimization algorithms have to grapple with. In the absence of any assumptions of convexity on the objective function, a global algorithm is forced to explore enough until it reaches a point in the process when with some degree of certainty it can localize its search space and perform local optimization (exploitation). Derivative bounds such as the ones discussed here together with the boundedness of the search space, guaranteed by the compactness assumption on $\mathcal{D}$, provide us with such certainty by producing a useful upper bound that allows us to shrink the search space. This is illustrated in Figure 1. Suppose we know that our function is Lipschitz with constant $L$, then given sample points as shown in the figure, we can use the Lipschitz property to discard pieces of the search space. This is done by finding points in the search space where the function could not possibly be higher than the maximum value already encountered. Such points are found by placing cones at the sampled points with slope equal to $L$ and checking where those cones lie below the maximum observed value.

This crude approach is wasteful because very often the slope of the function is much smaller than $L$. As shown in Figure 2), GPs do a better job of providing lower and upper bounds that can be used to limit the search space, by essentially choosing Lipschitz constants that vary over the search space and the algorithm run time.

We also assume that the objective function $f$ is costly to evaluate. We would like to avoid probing $f$ as much as possible and to get close to the optimum as quickly as possible. A solution to this problem is to approximate $f$ with a *surrogate function* that provides a good upper bound for $f$ and which is easier to calculate and optimize (Brochu et al., 2009). Surrogate functions also aid with global optimization by restricting the domain of interest. The surrogate that we will make extensive use of here is called the Upper Confidence Bound (UCB). It is defined to be $\mu + B\sigma$, where $\mu$ and $\sigma$ are the posterior predictive mean and standard deviation of the GP and $B$ is a constant to be chosen by the algorithm. This surrogate function has been studied extensively in the literature and this paper relies heavily on the ideas put forth in the paper by Srinivas et al (Srinivas et al., 2010), in which the algorithm consists of repeated optimization of the UCB surrogate function after each sample. It must be noted however that our algorithm is distinctly different from their UCB algorithm.

One key difference between our setting and that of (Srinivas et al., 2010) is that, whereas we assume that the value of the function can be observed exactly, for the analysis presented in (Srinivas et al., 2010) it is necessary for the noise to be non-trivial (and Gaussian) because the main quantity that is used in the estimates, namely information gain, cf. (Srinivas et al., 2010, Equation 3), becomes undefined when the variance of the observation noise ($\sigma^2$ in their notation) is set to 0, cf. the expression for $I(\mathbf{y}_A; \mathbf{f}_A)$ that was given in the paragraph following Equation (3). So, our analysis is complementary to theirs. Of course, one could still use their algorithm in the noiseless setting, but their analytical results are inapplicable to that case. Moreover, we show that the regret, $r(x_t) = \max_{\mathcal{D}} f - f(x_t)$, decreases according to $\mathcal{O}\left(e^{-\frac{\tau t}{(\ln t)^{d/4}}}\right)$, implying that the cumulative regret is bounded from above.

The paper whose results are most similar to ours is (Munos, 2011), but there are some key differences in the methodology, analysis and obtained rates. For instance, we are interested in cumulative regret, whereas the results of (Munos, 2011) are proven for finite stop-time regret. In our case, the ideal application is the optimization of a function that is $C^2$-smooth and has an unknown non-singular Hessian at the maximum. We obtain a regret rate $\mathcal{O}\left(e^{-\frac{\tau t}{(\ln t)^{d/4}}}\right)$, whereas the DOO algorithm in (Munos, 2011) has regret rate $\mathcal{O}(e^{-t})$ if the Hessian is known and the SOO algorithm has regret rate $\mathcal{O}(e^{-\sqrt{t}})$ if the Hessian is unknown. In addition, the algorithms in (Munos, 2011) can handle functions that behave like $-c\|x - x_M\|^\alpha$ near the maximum (cf. Example 2 therein). Moreover, the hierarchical decomposition of the search space utilized by DOO and SOO makes them much more efficient in practice than the algorithm presented in this paper: this is a shortcoming of our algorithm that we would like to remedy in the future.

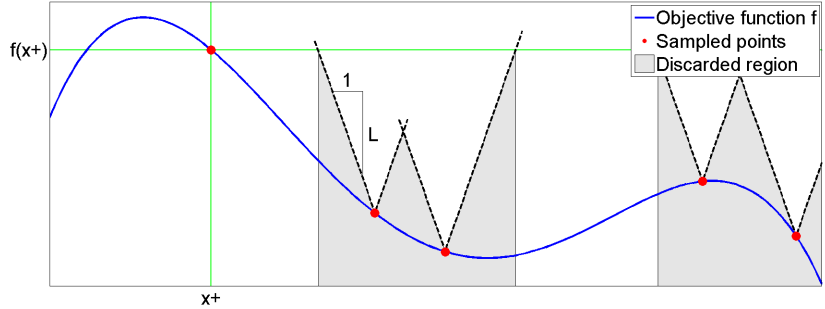

Figure 1. An example of the Lipschitz hypothesis being used to discard pieces of the search space when finding the maximum of a function $f$. Although $f$ is only known at the red sample points, if the derivative upper bounds (dashed lines) are below the best attained value thus far, $f(x^+)$, the corresponding areas of the search space (shaded regions) may be discarded.

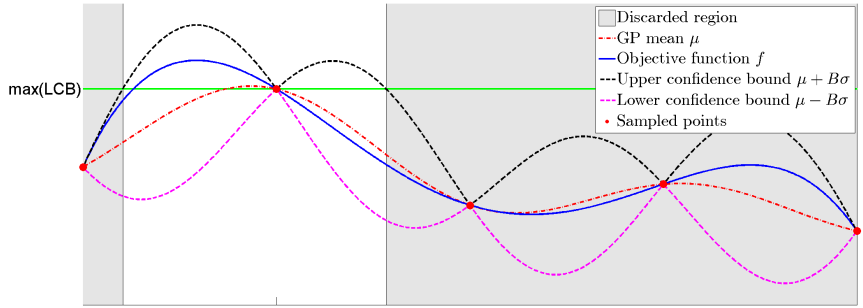

Figure 2. An example of our branch and bound maximization algorithm with UCB surrogate $\mu + B\sigma$, where $\mu$ and $\sigma$ are the mean and standard deviation of the GP respectively. The region consisting of the points $x$ for which the upper confidence bound $\mu(x) + B\sigma(x)$ is lower that the maximum value of the lower confidence bound $\mu(x) - B\sigma(x)$ does not need to be sampled anymore. Note that the UCB surrogate function bounds $f$ from above.

This problem was also studied by (Vazquez & Bect, 2010) and (Bull, 2011), but using the Expected Improvement surrogate instead of UCB. Our methodology and results are different, but complementary to theirs.

## 2. Gaussian process bandits

### 2.1. Gaussian processes

As in (Srinivas et al., 2010), the objective function is distributed according to a Gaussian process prior:

$$f(x) \sim \mathrm{GP}(m(\cdot), \kappa(\cdot, \cdot)). \quad (1)$$

For convenience, and without loss of generality, we assume that the prior mean vanishes, i.e., $m(\cdot) = 0$. There are many possible choices for the covariance kernel. One obvious choice is the anisotropic kernel $\kappa$ with a vector of known hyperparameters (Rasmussen & Williams, 2006):

$$\kappa(x_i, x_j) = \widetilde{\kappa}\left(-(x_i - x_j)^\top \mathbf{D}(x_i - x_j)\right), \quad (2)$$

where $\widetilde{\kappa}$ is an isotropic kernel and $\mathbf{D}$ is a diagonal matrix with positive hyperparameters along the diagonal and zeros elsewhere. Our results apply to squared exponential kernels and Matérn kernels with parameter $\nu \geq 2$. In this paper, we assume that the hyperparameters are fixed and known in advance.

We can sample the GP at $t$ points by choosing points $\mathbf{x}_{1:t} := \{x_1, \ldots, x_t\}$ and sampling the values of the function at these points to produce the vector $\mathbf{f}_{1:t} = [f(x_1) \cdots f(x_t)]^\top$. The function values are distributed according to a multivariate Gaussian distribution $\mathcal{N}(0, \mathbf{K})$, with covariance entries $\kappa(x_i, x_j)$. Assume that we already have several observations from previous steps, and that we want to decide what action $x_{t+1}$ should be considered next. Let us denote the value of the function at this arbitrary new point as $f_{t+1}$. Then, by the properties of GPs,

$\mathbf{f}_{1:t}$ and $f_{t+1}$ are jointly Gaussian:

$$\begin{bmatrix} \mathbf{f}_{1:t} \\ f_{t+1} \end{bmatrix} \sim \mathcal{N}\left(\mathbf{0}, \begin{bmatrix} \mathbf{K} & \mathbf{k}^\top \\ \mathbf{k} & \kappa(x_{t+1}, x_{t+1}) \end{bmatrix}\right),$$

where $\mathbf{k} = [\kappa(x_{t+1}, x_1) \cdots \kappa(x_{t+1}, x_t)]^\top$. Using the Schur complement, one arrives at an expression for the posterior predictive distribution:

$$P(f_{t+1}|\mathbf{x}_{1:t+1}, \mathbf{f}_{1:t}) = \mathcal{N}(\mu_t(x_{t+1}), \sigma_t^2(x_{t+1})),$$

where

$$\begin{aligned} \mu_t(x_{t+1}) &= \mathbf{k}^\top \mathbf{K}^{-1} \mathbf{f}_{1:t}, \\ \sigma_t^2(x_{t+1}) &= \kappa(x_{t+1}, x_{t+1}) - \mathbf{k}^\top \mathbf{K}^{-1} \mathbf{k} \end{aligned} \quad (3)$$

and $\mathbf{f}_{1:t} = [f(x_1) \cdots f(x_t)]^\top$.

### 2.2. Surrogates for optimization

When it is assumed that the objective function $f$ is sampled from a GP, one can use a combination of the posterior predictive mean and variance given by Equations (3) to construct surrogate functions, which tell us where to sample next. Here we use the UCB combination, which is given by

$$\mu_t(x) + B_t \sigma_t(x),$$

where $\{B_t\}_{t=1}^\infty$ is a sequence of numbers specified by the algorithm. This surrogate trades-off exploration and exploitation since it is optimized by choosing points where the mean is high (exploitation) and where the variance is large (exploration). Since the surrogate has an analytical expression that is easy to evaluate, it is much easier to optimize than the original objective function. Other popular surrogate functions constructed using the sufficient statistics of the GP include the Probability of Improvement, Expected Improvement and Thompson sampling. We refer the reader to (Brochu et al., 2009; May et al., 2010; Hoffman et al., 2011) for details on these.

### 2.3. Our algorithm

The main idea of our algorithm (Algorithm 1) is to tighten the bound on $f$ given by the UCB surrogate function by sampling the search space more and more densely and shrinking this space as more and more of the UCB surrogate function is "submerged" under the maximum of the Lower Confidence Bound (LCB). Figure 2 illustrates this intuition.

More specifically, the algorithm consists of two iterative stages. During the first stage, the function is sampled at enough points in $\mathcal{L}$ (the red crosses in Figure 3) until every point in the search space is contained inside a simplex of diameter $\delta$, where by the diameter of a set we mean the maximum length between any pair of points in the set. In the second stage, the search space is shrunk to discard regions where the maximum is very unlikely to reside. Such regions are obtained by finding points where the UCB is lower than the LCB (the complement of the colored region in the same panel as before). The remaining set of relevant points is denoted by $\widetilde{\mathcal{R}}$. To simplify the task of shrinking the search space, we simply find an enclosing ball, which is denoted by $\mathcal{R}$ in Algorithm 1. Back to the first stage, we consider a lattice that is twice as dense as in the first stage of the previous iteration, but we only sample at points that lie within our new smaller search space.

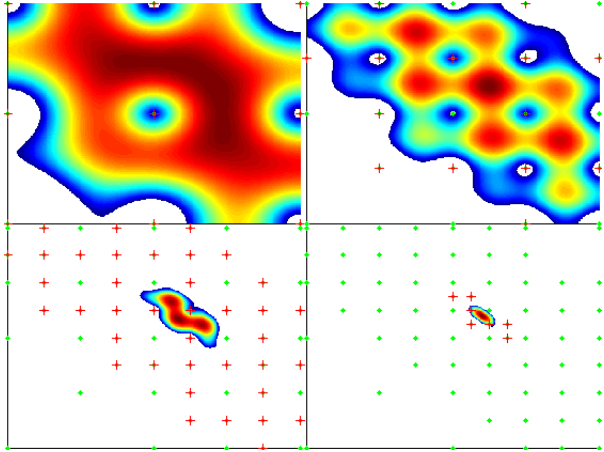

Figure 3. Branch and Bound algorithm for a 2D function. The colored region is the search space and the colormap, with red high and blue low, illustrates the value of the UCB. Four steps of the algorithm are shown; progressing from left to right and top to bottom. The green dots designate the points where the function was sampled in the previous steps, while the red crosses denote the freshly sampled points.

In the second stage, the auxiliary step of approximating the relevant set $\widetilde{\mathcal{R}}$ with the ball $\mathcal{R}$ introduces inefficiencies in the algorithm, since we only need to sample inside $\widetilde{\mathcal{R}}$. This can be easily remedied in practice to obtain an efficient algorithm. Our analysis will show that even without these improvements it is already possible to obtain very strong exponential convergence rates. Of course, practical improvement will result in better constants and ought to be considered seriously.

Note that Algorithm 1 terminates once the relevant region becomes too small to intersect the lattice $\mathcal{L}$. Our analysis requires for the algorithm to sample

**Algorithm 1** Branch and Bound

Input: A compact subset $\mathcal{D} \subseteq \mathbb{R}^d$, a function $f : \mathcal{D} \to \mathbb{R}$ and a discrete lattice $\mathcal{L} \subseteq \mathcal{D}$ that is divisible by powers of 2. Set $\mathcal{R} \leftarrow \mathcal{D}$ and $\delta \leftarrow 1$.
**repeat**
  **Sample Twice as Densely:**
    • $\delta \leftarrow \delta/2$.
    • Sample $f$ at enough points in $\mathcal{L}$ so that every point in $\mathcal{R}$ is contained in a simplex of diameter $\delta$.
  **Shrink the Relevant Region:**
    • Set
    $$\widetilde{\mathcal{R}} := \left\{ x \in \mathcal{R} \,\middle|\, \mu_T(x) + \sqrt{\beta_T}\sigma_T(x) > \sup_{\mathcal{R}} \mu_T(x) - \sqrt{\beta_T}\sigma_T(x) \right\}.$$

    $T$ is the number points sampled so far and $\beta_T = 2 \ln\left(\frac{|\mathcal{L}|T^2}{\alpha}\right) = 4\ln T + 2\ln\frac{|\mathcal{L}|}{\alpha}$ with $\alpha \in (0,1)$.
    • Solve the following constrained optimization problem: $(x_1^*, x_2^*) = \operatorname{argsup}_{(x_1,x_2) \in \widetilde{\mathcal{R}} \times \widetilde{\mathcal{R}}} \|x_1 - x_2\|$.
    • $\mathcal{R} \leftarrow B\left(\dfrac{x_1^* + x_2^*}{2}, \|x_1^* - x_2^*\|\right)$, where $B(p, r)$ is the ball of radius $r$ centred around $p$.
**until** $\mathcal{R} \cap \mathcal{L} = \varnothing$

---

points from a fixed finite set of points, although we can pick $\mathcal{L}$ to be the set of all points in $\mathcal{D}$ with floating point coordinates.

## 3. Analysis

We begin our analysis by showing that, given sufficient explored locations, the posterior predictive variance is small. Specifically, the following approximation result is proved in the supplementary material:

**Proposition 1 (Variance Bound)** *Let $\kappa : \mathbb{R}^d \times \mathbb{R}^d \to \mathbb{R}$ be a kernel that is four times differentiable along the diagonal $\{(x, x) \,|\, x \in \mathbb{R}^d\}$, with $Q$ defined as in part 2 of Lemma 5, and $f \sim \operatorname{GP}(0, \kappa(\cdot, \cdot))$ a sample from the corresponding GP. If $f$ is sampled at points $x_{1:T} = \{x_1, \ldots, x_T\}$ that form a $\delta$-cover of a subset $\mathcal{D} \subseteq \mathbb{R}^d$, then the resulting posterior predictive standard deviation $\sigma_T$ satisfies*

$$\sup_{\mathcal{D}} \sigma_T \leq \frac{Q\delta^2}{4}.$$

### 3.1. Finiteness of regret

Having shown that the variance vanishes according to the square of the resolution of the lattice of sampled points, we now move on to show that this estimate implies an exponential asymptotic vanishing of the regret encountered by our Branch and Bound algorithm. This is laid out in our main theorem stated below and proven in the supplementary material.

Recall that $\mathcal{D} \subseteq \mathbb{R}^d$ is assumed to be a nonempty compact subset and $f$ a sample from the GP $\operatorname{GP}(0, \kappa(\cdot, \cdot))$ on $\mathcal{D}$. Moreover, in what follows we will denote the global maximum by $x_M := \operatorname{argmax}_{x \in \mathcal{D}} f(x)$ and the regret by $r(x_t) = f(x_M) - f(x_t)$. Also, by convention, for any set $\mathcal{S}$, we will denote its interior by $\mathcal{S}^\circ$, its boundary by $\partial \mathcal{S}$ and if $S$ is a subset of $\mathbb{R}^d$, then $\operatorname{conv}(S)$ will denote its convex hull. The following holds true:

**Theorem 2** *Suppose we are given:*
1. *$\alpha > 0$, a compact subset $\mathcal{D} \subseteq \mathbb{R}^d$, and $\kappa$ a kernel on $\mathbb{R}^d$ that is four times differentiable along the diagonal;*
2. *$f \sim \operatorname{GP}(0, \kappa)$ a continuous sample on $\mathcal{D}$ that has a unique global maximum $x_M$, which satisfies one of the following two conditions:*
   (†) *$x_M \in \mathcal{D}^\circ$ and $f(x_M) - c_1\|x - x_M\|^2 < f(x) \leq f(x_M) - c_2\|x - x_M\|^2$ for all $x$ satisfying $x \in B(x_M, \rho_0)$ for some $\rho_0 > 0$;*
   (‡) *$x_M \in \partial\mathcal{D}$ and both $f$ and $\partial\mathcal{D}$ are smooth at $x_M$, with $\nabla f(x_M) \neq 0$;*
3. *any lattice $\mathcal{L} \subseteq \mathcal{D}$ satisfying the following two conditions*
   • $2\mathcal{L} \cap \operatorname{conv}(\mathcal{L}) \subseteq \mathcal{L}$ \hfill (4)
   • $2^{\lceil -\log_2 \frac{\rho_0}{\operatorname{diam}(\mathcal{D})}\rceil + 1}\mathcal{L} \cap \mathcal{L} \neq \varnothing$ \hfill (5)
   *if $f$ satisfies* (†)

*Then, there exist positive numbers $A$ and $\tau$ and an integer $T$ such that the points specified by the Branch and Bound algorithm, $\{x_t\}$, will satisfy the following asymptotic bound: For all $t > T$, with probability $1 - \alpha$ we have*

$$r(x_t) < A e^{-\frac{\tau t}{(\ln t)^{d/4}}}.$$

Given the exponential rate of convergence we obtain in Theorem 2, we have the following finiteness conclusion for the cumulative regret accrued by our Branch and Bound algorithm:

**Corollary 3** *Given $\kappa$, $f \sim \mathrm{GP}(0, \kappa)$ and $\mathcal{L} \subseteq \mathcal{D}$ as in Theorem 2, the cumulative regret is bounded from above.*

**Remark 4** *It is worth pointing out the trivial observation that using a simple UCB algorithm with monotonically increasing and unbounded factor $\sqrt{\beta_t}$, without any shrinking of the search space as we do here, necessarily leads to unbounded cumulative regret since eventually $\sqrt{\beta_t}$ becomes large enough so that at points $x'$ far away from the maximum, $\sqrt{\beta_t}\sigma_t(x')$ becomes larger than $f(x_M) - f(x)$. In fact, eventually the UCB algorithm will sample every point in the lattice $\mathcal{L}$.*

### 3.2. Remarks on the main theorem

This section includes a discussion of the assumptions placed on the objective function in Theorem 2 as well as an outline of the proof, the full details of which are included in the appendix.

#### 3.2.1. ON THE STATEMENT OF THEOREM 2

A few remarks on the assumptions and the conclusion of the main theorem are in order:

**A. Relationship between the local and global assumptions on $f$:** The theorem has two seemingly unrelated restrictions on the function $f$: the global GP prior and the local behaviour near the global maximum. However, in many circumstances of interest, the local condition follows almost surely from the global condition. Two such circumstances are if $\kappa$ is a Matérn kernel with $\nu > 2$ (including the squared exponential kernel) or if $\kappa$ is six times differentiable. In either case, the sample $f$ is twice differentiable almost surely, in the former case by (Adler & Taylor, 2007, Theorem 1.4.2) and (Stein, 1999, §2.6)) and in the latter situation by (Ghosal & Roy, 2006, Theorem 5). If the global maximum $x_M$ lies in the interior of $\mathcal{D}$, the Hessian of $f$ at $x_M$ will almost surely be non-singular since the vanishing of at least one of the eigenvalues of the Hessian is a codimension 1 condition in the space of all functions that are smooth at a given point, hence justifying condition (†).

On the other hand, if $x_M$ lies on the boundary of $\mathcal{D}$, then condition (‡) will be satisfied almost surely, since the additional event of the vanishing of $\nabla f(x_M)$ is a codimension $d$ phenomenon in the space of functions with global maximum at $x_M$.

**B. Uniqueness of the global maximum:** A randomly drawn continuous sample from a GP on a compact domain will almost surely have a unique global maximum: this is because the space of continuous functions on a compact domain that attain their global maximum at more than one point have codimension one in the space of all continuous functions on that domain.

**C. Assumptions on $\mathcal{L}$:** The two conditions (4) and (5) simply require that the lattice be "divisible by 2" and that it be fine enough so that the algorithm can sample inside the ball $B(x_M, \rho_0)$ when the maximum of the function is located in the interior of the search space $\mathcal{D}$. One can simply choose $\mathcal{L}$ to be the set of points in $\mathcal{D}$ that have floating point coordinates: it's just the points at which the algorithm is allowed to sample the function.

**D. On $\tau$'s dependence:** Finally, it is important to point out that the decay rate $\tau$ does not depend on the choice of the lattice $\mathcal{L}$, even though as stated, the statement of the theorem chooses $\tau$ only after $\mathcal{L}$ is specified. The theorem was written this way simply for the sake of readability.

#### 3.2.2. OUTLINE OF THE PROOF OF THEOREM 2

The starting point for the proof is the observation that one can use the posterior predictive mean and standard deviation of the GP to obtain a high probability envelope around the objective function (cf. Lemma 8 in the appendix). Given the fact that the thickness of this envelope is determined by the height of the posterior predictive standard deviation, $\sigma$, we can use the bound given by Proposition 1 to show that asymptotically one can rapidly dispense with large portions of the search space, as illustrated in Figure 2.

One disconcerting component of Algorithm 1 is the step that requires sampling twice as densely in each iteration, since the number of samples can start to grow exponentially, hence killing any hope of obtaining exponentially decreasing regret. However, this is where the assumption on the local behaviour near the global maximum becomes relevant. Since Proposition 1 tells us that every time the function is sampled twice as densely, $\sigma$ decreases by a factor of 4, and given our assumption that the function has quadratic behaviour near the global maximum, we can conclude that the radius of the search space is halved after each iteration and so the number of sam-pled points added in each iteration roughly remains constant. Of course, this assumes that the multiplicative factor $\sqrt{\beta_t}$ remains constant in this process. However, the algorithm requires $\sqrt{\beta_t}$ to grow logarithmically, and so to fill this gap we need to bound the growth of $\sqrt{\beta_t}$, which is tied to the number of samples needed in each iteration of the algorithm, which in turn is linked to the resolution of the lattice of sampled points $\delta$ and the size of the relevant set $\mathcal{R}$, which in turn depends on the size of $\sqrt{\beta_t}\sigma_t$. This circular dependence gives rise to a difference equation, whose solutions we bound by solving the corresponding differential equation.

### 3.2.3. FURTHER REMARKS ON THE GP PRIOR

Let us step back for a moment and pose the question of whether it would be possible to carry out a similar line of reasoning under other circumstances. To answer this, one needs to identify the key ingredients of the proof, which are the following:

A. A mechanism for calculating a high probability envelope around the objective function (cf. Lemma 8);

B. An estimate showing that the thickness of the envelope diminishes rapidly as the function is sampled more and more densely (cf. Proposition 1), so that the search space can be shrunk under reasonable assumptions on the behaviour of the function near the peak.

The reason for our imposing a GP prior on $f$ is that it gives us property A, while our smoothness assumption on the kernel guarantees property B. However, GPs are but one way one could obtain these properties and they do this essentially by coming up with local estimates of the Lipschitz constant based on the observed values of the objective function nearby. Perhaps one could explicitly incorporate similar local estimates on the Lipschitz constant into tree based approaches like HOO and SOO, cf. (Bubeck et al., 2011) and (Munos, 2011), in which case one would be able to dispense with the GP assumption and get similar performance. But, that is beyond the scope of this paper and will be left for future work.

## 4. Discussion

In this paper we proposed a modification of the UCB algorithm of (Srinivas et al., 2010) which addresses the noise free case. The key difference is that while the original algorithm achieves an $O(t^{-\frac{1}{2}})$ rate of convergence to the regret minimizer, we obtain an exponential rate in the number of function evaluations. In other words, the noise free problem is significantly easier, statistically speaking, than the noisy case. The key difference is that we need not invest any samples in noise reduction to determine whether our observations deviate far from their expectation.

This allows us to discard pieces of the search space where the maximum is very unlikely to be, when compared to (Srinivas et al., 2010). We show that this additional step leads to a considerable improvement of the regret accrued by the algorithm. In particular, the cumulative regret obtained by our Branch and Bound algorithm is bounded from above, whereas the cumulative regret bound obtained in the noisy bandit algorithm is unbounded. The possibility of dispensing with chunks of the search space can also be seen in the works involving hierarchical partitioning, e.g. (Munos, 2011), where regions of the space are deemed as less worthy of probing as time goes on.

Our results mirror the observation in active learning that noise free and large margin learning of half spaces can be achieved much more rapidly than identifying a linear separator in the noisy case (Bshouty & Wattad, 2006; Dasgupta et al., 2009). This is also reflected in classical uniform convergence results for supervised learning (Audibert & Tsybakov, 2007; Vapnik, 1998) where the achievable rate depends on the decay of probability mass near the margin.

This suggests that the ability to extend our results to the noisy case is somewhat limited. An indication of what might be possible can be found in (Balcan et al., 2009), where regions of the version space are eliminated once they can be excluded with sufficiently high probability. One could model a corresponding Branch and Bound algorithm, which dispenses with points that lie outside the current (or perhaps the previous) relevant set when calculating the covariance matrix $\mathbf{K}$ in the posterior equations (3). Analysis of how much of an effect such a computational cost-cutting measure would have on the regret encountered by the algorithm is a subject of future research.

## Acknowledgements

We are very grateful to the anonymous reviewers for outstanding feedback. This research was supported by NSERC and the Institute for Computing, Information and Cognitive Systems (ICICS) at UBC.